\documentclass{article}

\usepackage[preprint]{neurips_2026}

\usepackage[utf8]{inputenc}
\usepackage[T1]{fontenc}
\usepackage{hyperref}
\usepackage{url}
\usepackage{booktabs}
\usepackage{amsfonts}
\usepackage{amsmath}
\usepackage{nicefrac}
\usepackage{microtype}
\usepackage{xcolor}
\usepackage{graphicx}
\usepackage{algorithm}
\usepackage{algorithmic}
\usepackage{natbib}
\usepackage{xcolor}
\usepackage{wrapfig}
\usepackage{multirow}

\title{Offline Semantic Guidance for Efficient Vision-Language-Action Policy Distillation}

\author{%
Jin Shi\thanks{Equal contribution.} \\
Department of Mechanical Engineering \\
University College London \\
London, United Kingdom \\
\texttt{
jin.shi.24@ucl.ac.uk}
\And
Brady Zhang\footnotemark[1] \\
Department of Mechanical Engineering \\
University College London \\
London, United Kingdom \\
\texttt{brady.zhang.24@ucl.ac.uk}
\And
Yishun Lu \\
Department of Engineering Science \\
University of Oxford \\
Oxford, United Kingdom \\
\texttt{yishun.lu@eng.ox.ac.uk}
}

\begin{document}

\maketitle

\begin{abstract}
Billion-parameter Vision-Language-Action (VLA) policies have recently shown impressive performance in robotic manipulation, yet their size and inference cost remain major obstacles for real-time closed-loop control. We introduce \textbf{VLA-AD}, a distillation framework that uses a Vision-Language Model as an offline semantic supervisor to transfer large VLA teachers into lightweight student policies. Instead of relying only on low-level action imitation, VLA-AD augments teacher-provided 7-DoF action targets with high-level semantic guidance, including task phase anchors and multi-frame operating-direction descriptions. These auxiliary signals are used only during training: at test time, the student policy runs independently, with neither the VLA teacher nor the VLM required.
We evaluate VLA-AD on three LIBERO benchmark suites. Using OpenVLA-7B as the teacher, our method produces a 158M-parameter student, yielding a $44\times$ reduction in model size while matching the teacher with only a $0.27\%$ average relative gap. The resulting policy runs at 12.5 Hz on an RTX 4090, achieving a $3.28\times$ inference speedup over OpenVLA-7B. We further show that the same semantic distillation pipeline generalizes to a different $\pi_{0.5}$-4B teacher, where the student outperforms the teacher on two suites and remains within $0.53\%$ on \texttt{libero\_goal}. Additional analysis indicates that phase-level supervision and multi-frame directional cues make the student less sensitive to noisy teacher actions, such as erroneous high-frequency gripper changes. Overall, VLA-AD demonstrates that offline semantic guidance from VLMs can substantially improve the efficiency, robustness, and deployability of VLA policy distillation.
\end{abstract}

\section{Introduction}

In recent years, robotic manipulation has undergone a paradigm shift with the emergence of large-scale vision-language-action (VLA) foundation models . By combining transformer architectures with large-scale vision-language pretraining and real-world robotic trajectories, these models have demonstrated strong multi-task generalization and zero-shot instruction-following capabilities ~\citep{brohan2023rt2visionlanguageactionmodelstransfer,kim2024openvlaopensourcevisionlanguageactionmodel}. They represent an important step toward generalist robot policies that can execute diverse language-conditioned manipulation tasks in complex, unstructured environments. Despite these advances, deploying billion-parameter VLA models on real robotic platforms remains highly challenging. A major bottleneck is inference latency \citep{ahn2024autortembodiedfoundationmodels}. Evaluating 7B-parameter or larger models on edge devices commonly used in robotics can require more than one second per control step \citep{brohan2023rt2visionlanguageactionmodelstransfer,lin2026awqactivationawareweightquantization}. A common solution is to distill these large models into compact policies using standard behavioral cloning (BC) ~\citep{ross2011reduction,mandlekar2021matterslearningofflinehuman}. However, this introduces a second challenge: vulnerability to compounding errors. Since single-task robotic datasets typically cover only a limited subset of possible states, cloned student policies can suffer from severe distribution shift during closed-loop execution ~\citep{chen2019learningcheating}. As a result, a student may achieve high offline action-prediction accuracy but fail completely during online evaluation. Small prediction errors can accumulate over hundreds of timesteps, driving the robot into out-of-distribution (OOD) states that were not observed during training.

Recent advances suggest that vision-language models (VLMs) can provide a useful source of auxiliary supervision for addressing this problem. Although VLMs do not directly output low-level 7-DoF robot actions, they possess strong scene understanding, spatial reasoning, and commonsense physical priors ~\citep{huang2023voxposercomposable3dvalue,bai2025qwen25vltechnicalreport}. Prior work has shown that aligning robotic policies with VLM-derived representations can improve semantic awareness and OOD generalization ~\citep{kachaev2025dontblindvlaaligning,karamcheti2023languagedrivenrepresentationlearningrobotics}. This motivates the use of a VLM as an offline ``knowledge translator'': it can extract task-relevant visual-semantic information from demonstrations and inject this information into a lightweight control policy, thereby bridging raw visual observations and high-level manipulation logic ~\citep{li2024visionlanguagefoundationmodelseffective,ahn2022icanisay}.

Motivated by this observation, we introduce \textbf{VLA-AD}, a VLM-assisted distillation framework for converting large VLA teachers into efficient closed-loop policies. VLA-AD uses a VLM as a parallel offline supervisor during training, while removing the VLM entirely at deployment. This allows the student to benefit from high-level semantic supervision without increasing inference latency. We evaluate VLA-AD through closed-loop experiments on the LIBERO benchmark and make the following contributions:

\begin{itemize}
    \item \textbf{VLM-Assisted Zero-Overhead Distillation.}
    We propose VLA-AD, a parallel-supervision distillation framework that uses a VLM as an offline semantic supervisor for compressing billion-scale VLA teachers into compact student policies. The VLM provides phase and direction descriptions during training, but is removed entirely at deployment, introducing zero additional inference latency. With OpenVLA-7B as the teacher, VLA-AD allows a 158M-parameter student with $44\times$ parameter compression, matching the teacher within only $0.27\%$ difference on average across three LIBERO suites. On an RTX~4090, the student reaches 12.5~Hz, giving a $3.28\times$ speedup over OpenVLA-7B.

\item \textbf{Teacher-Agnostic Semantic Supervision Beyond Teacher Performance.}
We show that VLA-AD is not tied to a specific teacher architecture: the same semantic-supervision pipeline transfers across both OpenVLA-7B and $\pi_{0.5}$-4B teachers. Despite using a compact 158M-parameter student, VLA-AD matches OpenVLA-7B within only $0.27\%$ on relative average across three LIBERO suites, and surpasses the $\pi_{0.5}$-4B teacher on two of the three suites while remaining within $0.53\%$ on the \texttt{libero\_goal}. This indicates that VLM-generated phase and direction anchors capture transferable manipulation structure, allowing the student to learn closed-loop behaviours that are not merely bounded by the raw teacher action distribution.

\item \textbf{Phase Anchors and Multi-Frame Direction for Robust Distillation.}
We introduce a structured semantic-anchor mechanism that combines a 9-class phase descriptor with a multi-frame operating-direction signal. The phase descriptor provides a stable abstraction of the current manipulation stage, while the multi-frame direction signal resolves temporal ambiguity in visually similar operating states. This design gives the student consistent spatio-temporal context, helping it infer manipulation intent rather than overfitting to isolated frame-level observations.

\end{itemize}

Overall, VLA-AD provides a practical route toward deploying compact and robust VLA policies on real robotic platforms. By transferring semantic knowledge only during training, it improves closed-loop robustness while preserving the low-latency inference required for real-time control.

\section{Related Work}
\subsection{Distillation of Large Robotic Policies}

Deploying large-scale vision-language-action (VLA) foundation models~\citep{kim2024openvlaopensourcevisionlanguageactionmodel,brohan2023rt2visionlanguageactionmodelstransfer} on edge devices requires substantial model compression. Recent work has therefore explored VLA distillation as a way to reduce inference latency. TinyVLA~\citep{wen2025tinyvlafastdataefficientvisionlanguageaction} focuses on data-efficient architectural miniaturisation for high-frequency control, while CEED-VLA~\citep{song2025ceedvlaconsistencyvisionlanguageactionmodel} improves efficiency through early-exit decoding and consistency distillation. Other methods, such as VITA-VLA~\citep{dong2025vitavlaefficientlyteachingvisionlanguage} and ActDistill~\citep{ye2026actdistillgeneralactionguidedselfderived}, aim to transfer manipulation priors using action-expert distillation or dynamic action-guided routing.

Despite these advances, standard behavioural cloning (BC) remains vulnerable to compounding errors and covariate shift during closed-loop execution~\citep{ross2011reduction,mandlekar2021matterslearningofflinehuman}. Since single-task demonstration datasets usually cover only a limited range of states, small prediction errors can quickly drive the student policy into out-of-distribution (OOD) regions. Recent methods partially address this issue, but often rely on auxiliary action experts~\citep{dong2025vitavlaefficientlyteachingvisionlanguage}, dynamic routing mechanisms~\citep{ye2026actdistillgeneralactionguidedselfderived}, or architectural changes that may complicate deployment~\citep{song2025ceedvlaconsistencyvisionlanguageactionmodel}. 

\subsection{Phase Anchors and Temporal Context in Manipulation}

Decomposing long-horizon manipulation tasks into discrete phases or semantic primitives is a well-established strategy for reducing control complexity~\citep{ahn2022icanisay,bharadhwaj2023roboagentgeneralizationefficiencyrobot}. Recent work has used large language models (LLMs) and vision-language models (VLMs) to generate high-level subgoals, task plans, or step-by-step instructions for guiding low-level robot policies~\citep{huang2023voxposercomposable3dvalue}. In parallel, imitation learning methods have explored temporal context and action smoothing to address noisy demonstrations and short-term inconsistencies. Representative examples include observation-history conditioning, action chunking in ACT~\citep{zhao2023learningfinegrainedbimanualmanipulation}, and denoising-based trajectory prediction in Diffusion Policy~\citep{chi2024diffusionpolicyvisuomotorpolicy}.

However, existing phase-based or hierarchical approaches often require multi-stage policy architectures or online queries to large models during execution, which conflicts with the low-latency requirements of edge deployment. Meanwhile, although action chunking and history-conditioned policies can reduce high-frequency action noise, they primarily rely on raw visual or proprioceptive histories and provide limited explicit semantic grounding. This makes lightweight students vulnerable to ambiguous or inconsistent teacher labels, such as spurious gripper oscillations observed in foundation VLA rollouts.

\section{Method}\label{sec:method}

\subsection{Overall Architecture}\label{sec:pipeline}

\begin{figure}[ht]
\centering
\includegraphics[width=\columnwidth]{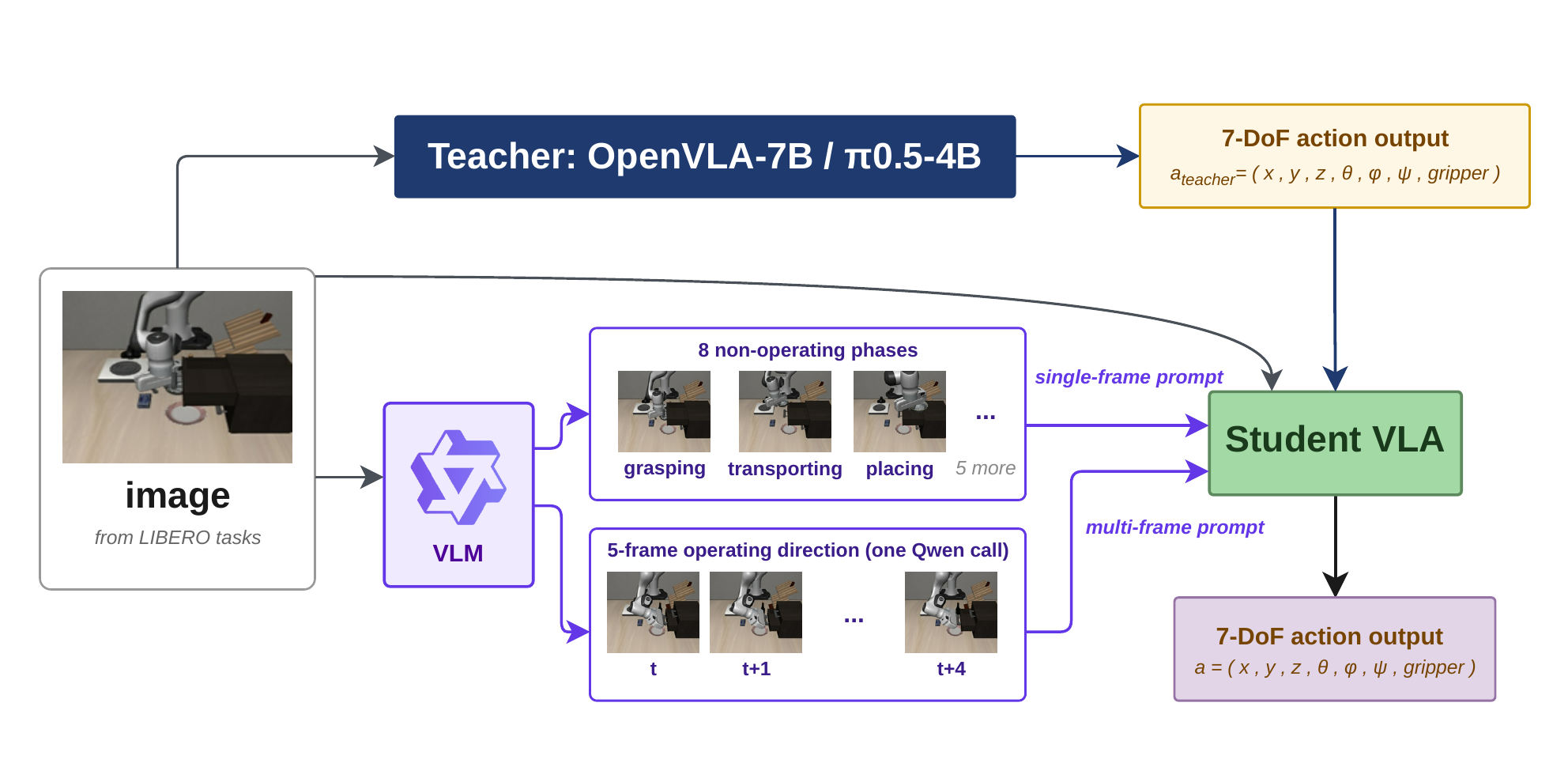}
\caption{VLA-AD: Overall VLM-Assisted Distillation pipeline. A lightweight Student VLA is trained via parallel offline supervision from a frozen teacher VLA (e.g., OpenVLA-7B or $\pi_{0.5}$-4B) and a Vision-Language Model (VLM). }
\label{fig:pipeline}
\end{figure}

We formulate VLA-AD as a parallel knowledge distillation framework designed to compress a computationally heavy VLA teacher (e.g., OpenVLA-7B or $\pi_{0.5}$-4B) into a highly efficient student VLA. As illustrated in Figure~\ref{fig:pipeline}, our offline training pipeline operates through a continuous, three-stage process. First, we collect expert manipulation trajectories by executing the teacher VLA across LIBERO tasks. We retain only the successful episodes to construct a curated dataset of raw visual observations paired with their corresponding 7-DoF expert actions, denoted as $a_{\text{teacher}} = (x, y, z, \theta, \phi, \psi, \text{gripper})$. Second, to bridge the representation gap between raw pixels and low-level continuous control, we leverage an off-the-shelf Vision-Language Model (VLM) to generate rich, phase-aware natural language annotations. This auxiliary semantic supervision is explicitly decoupled into two complementary streams: a single-frame prompt that directs the VLM to classify the current state into one of eight distinct non-operating phases (e.g., \textit{grasping}, \textit{transporting}), and a multi-frame prompt that queries the VLM across a 5-frame temporal window to extract the ongoing operating direction. Finally, the compact student VLA is trained end-to-end. It consumes the raw image observation alongside the VLM-generated single-frame and multi-frame text prompts as conditioned inputs, and is supervised directly by the teacher's 7-DoF action targets. By establishing this parallel supervision mechanism, the student effectively grounds its continuous action predictions within robust, human-interpretable semantic anchors.

\subsection{Motivation for Dual-Path Supervision}\label{sec:dual_path}

At each timestep, the student is required to output a 7-DoF action vector (comprising 3D translation increments ($\Delta x, \Delta y, \Delta z$), 3D rotation increments ($\Delta r_x, \Delta r_y, \Delta r_z$), and a 1D gripper state) while simultaneously maintaining a high-level understanding of the ongoing task phase. However, neither of our off-the-shelf supervisory sources provides a complete picture independently. On one hand, the teacher VLA (e.g., OpenVLA-7B or $\pi_{0.5}$-4B) supplies precise, per-frame 7-DoF action values, dictating exactly how much to move and whether to actuate the gripper. Yet, this continuous supervision lacks semantic grounding; the student receives numerical targets but remains unaware of the overarching task phase or the logical rationale behind discrete events, such as a sudden gripper closure. On the other hand, the VLM (Qwen2.5-VL) provides rich natural language descriptions that characterize the gripper state, relative object positions, and the current phase. However, the VLM operates purely in the semantic domain and cannot predict the low-level continuous action signals required for control.

In essence, the teacher demonstrates how to execute the manipulation, while the VLM explains what the current action is achieving. The student must synthesize both modalities during training so that it can independently generate temporally coherent and numerically accurate 7-DoF actions during evaluation (when both the teacher and VLM are removed). To achieve this, we introduce a dual-path training scheme. Both paths are supervised directly by the teacher's action targets. Formally, let $d_t$ denote the VLM-generated description for frame $t$, and $a_t^* \in \{0,\ldots,255\}^7$ the per-dimension discretized teacher action. The total dual-path loss is defined as:
\begin{equation}
\mathcal{L}_{\text{total}} \;=\; \mathcal{L}_{\text{full}}(x_t, \tau, d_t) \;+\; \alpha \cdot \mathcal{L}_{\text{img}}(x_t, \tau),
\label{eq:loss_total}
\end{equation}
where each per-path loss is a dimension- and chunk-weighted cross-entropy defined over a prediction horizon of $K$ steps:
\begin{equation}
\mathcal{L}_{\bullet} \;=\; \sum_{k=1}^{K}\sum_{j=1}^{7} w_j \cdot \text{CE}\!\big(\pi_S^{(k,j)}(\cdot \mid \bullet),\, a^{*}_{t+k,j}\big).
\label{eq:loss_path}
\end{equation}
Here, the weights $w = (1, 1, 1, 2, 2, 2, 1)$ are introduced to compensate for the smaller magnitude of the rotation channel. The full path, $\mathcal{L}_{\text{full}}$, utilizes the visual observation ($x_t$), task instruction ($\tau$), and semantic description ($d_t$). Conversely, the image-only path, $\mathcal{L}_{\text{img}}$, masks the description channel. This masking forces the visual representation to remain self-sufficient and prevents the student from collapsing into a description-only shortcut. The hyperparameter $\alpha$ dictates the weight of this auxiliary path. 

\subsection{Phase-Anchored Visual Description}\label{sec:phase}

Allowing a VLM like Qwen2.5-VL to freely describe the visual scene (as explored in our early pilot studies) introduces significant linguistic stochasticity. For the exact same manipulation stage, the VLM might output vastly different phrasing, such as ``the robot is grabbing'' versus ``reaching for.'' This descriptive noise is severely amplified when distilling into a compact 158M-parameter student. Unlike the 7B-parameter teacher, the lightweight student lacks the redundant capacity to absorb such linguistic variance. Consequently, representation-level inconsistencies directly corrupt the language conditioning, degrading the overall distillation quality.

To suppress the inherent randomness of the VLM, we introduce \textbf{phase-Anchored Visual Description}. Specifically, we employ a heuristic rule-based classifier (utilizing signals such as gripper state, 3D velocity, and task progression) to assign a fixed phase label to each frame. This phase label, along with its semantic definition, is then injected into the VLM's prompt. This constraint forces Qwen to explicitly use the assigned phase vocabulary while describing the relative object positions and the ongoing action. As a result, the description for every frame is anchored to a consistent semantic coordinate, leaving the VLM only the degree of freedom to clearly articulate the scene context in two to four natural language sentences.

\begin{figure}[ht]
\centering
\includegraphics[width=\columnwidth]{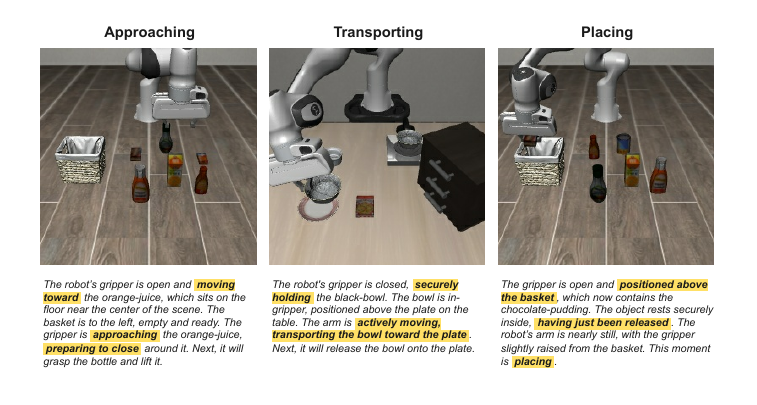}
\caption{Qualitative examples of Qwen2.5-VL phase-anchored descriptions on three OpenVLA-7B rollouts from \texttt{libero\_object}. Each panel shows a single frame together with the description Qwen produces under phase-anchored prompt for the phases \emph{approaching}, \emph{transporting}, and \emph{placing}. Highlighted text marks the action-semantic vocabulary that is tied to the phase label. }
\label{fig:phase_examples}
\end{figure}

During this phase-anchored annotation process, we inadvertently identified several anomalous behaviors exhibited by the teacher policy, likely stemming from noise in its original training distribution. As prominently illustrated in Figure~\ref{fig:gripper}, the OpenVLA-7B teacher frequently exhibits spurious, high-frequency gripper oscillations, emitting contradictory open/close signals even when the visual scene remains nearly identical. A standard behavioral cloning student would blindly overfit to these self-revoking labels. However, our phase classifier categorizes these anomalous sequences into stable, continuous phases (e.g., \textit{holding}). By combining these stable phase anchors with the semantic VLM descriptions, the student learns to prioritize the broader manipulation context over the noisy, frame-level action targets. Consequently, rather than inheriting the teacher's instability, the lightweight student becomes robust enough to smooth out these expert errors, outputting a significantly more consistent and logical control policy.

The number of distinct phases is a critical hyperparameter. We evaluated six different taxonomy granularities (3, 5, 7, 9, 11, and 13 phases) by measuring the Coefficient of Variation (CV) to assess how evenly the phases were distributed across the six LIBERO suites. The 9-phase taxonomy achieved the lowest deviation ($|CV - 1| = 0.14$), most closely approximating a Gaussian-like distribution. It strikes an optimal balance: it is neither too coarse to conflate distinct manipulation stages, nor too fine to result in long-tailed, undersampled phases. Consequently, we adopted a 9-phase taxonomy: \textit{idle}, \textit{approaching}, \textit{grasping}, \textit{transporting}, \textit{holding}, \textit{placing}, \textit{operating}, \textit{regrasping}, and \textit{completed}.

It is important to note that the specific lexical choices (e.g., ``grasping'' versus ``transporting'') are secondary. The student is not merely learning these isolated vocabulary words; rather, it is learning the comprehensive semantic representation that accompanies each phase label. This includes the VLM's full description of the gripper's current action, the relative spatial coordinates of the target object, spatial constraints, and the anticipated next step. The phase labels serve primarily as a linguistic anchor to standardize these rich, context-aware descriptions.




\subsection{Multi-Frame Operating Direction}\label{sec:direction}

While single-frame Qwen annotations suffice for the majority of manipulation phases, the \textit{operating} phase (e.g., opening a drawer or turning a knob) presents a unique challenge: directional ambiguity. From a single static frame, the direction of motion is often indiscernible. For example, a partially open drawer appears identical whether it is actively being pulled open or pushed closed; the spatial coordinates of the drawer, the robot's proprioceptive posture, and the gripper state are visually indistinguishable. This ambiguity introduces confounding noise into the student's learned policy. Because the supervision signal for the gripper channel must also implicitly capture the precise timing of an object's release, a lack of clear directional context can severely perturb the student's action predictions.

To resolve this, we introduce a multi-frame operating direction signal, formulated as a tuple: $(\text{element}, \text{direction})$. The $\text{element}$ specifies the targeted object or part being manipulated (e.g., \textit{drawer-handle}, \textit{knob}, \textit{bowl}), while the $\text{direction}$ characterizes its dynamic motion (e.g., \textit{outward}, \textit{clockwise}, \textit{downward}). 

While the other eight phases continue to rely exclusively on single-frame annotations, the \textit{operating} phase triggers a broader temporal context window. When a continuous sequence of frames is classified as \textit{operating}, we extract the entire segment and uniformly sample five keyframes spanning its beginning, middle, and end. These five frames are fed jointly to the VLM. By observing the visual progression across this temporal span, Qwen accurately infers the underlying $(\text{element}, \text{direction})$ tuple. This resulting tuple is then broadcast to every frame within that specific operating segment. For these frames, the final text description fed to the student is constructed by concatenating the phase label with the multi-frame tuple, providing a temporally stable and disambiguated semantic anchor.

\section{Experiments}\label{sec:experiments}

\subsection{Experimental Setup}
\label{sec:setup}

We evaluate VLA-AD on three LIBERO suites:
\texttt{libero\_object}, \texttt{libero\_spatial}, and
\texttt{libero\_goal} ~\citep{liu2023liberobenchmarkingknowledgetransfer}. Following the standard OpenVLA protocol, we conduct
20 closed-loop episodes across 10 tasks per suite, yielding 600 total
evaluation trials with a maximum horizon of 520 steps.
To demonstrate that our distillation pipeline is teacher-agnostic, we use two
expert models, OpenVLA-7B and $\pi_{0.5}$-4B, to supervise a unified
158M-parameter student policy. The student architecture comprises a Long-CLIP encoder ~\citep{zhang2024longclipunlockinglongtextcapability}, LoRA adapters ~\citep{hu2021loralowrankadaptationlarge}, and a tri-stream MLP head that predicts an action chunk
of length $K=5$ ~\citep{zhang2025mixtureexpertslargelanguage}. This achieves $44\times$ and $25\times$ parameter compression
relative to the respective teachers. To ensure a fair comparison, we re-evaluate
both teachers under our exact closed-loop protocol rather than relying on
officially reported metrics.
For each teacher, we compare a \textit{no-VLM} baseline, which is conditioned
solely on raw images and instructions, against our proposed \textit{with-VLM}
method, which integrates phase descriptions and multi-frame directions. For the
latter, we sweep the dual-path hyperparameter
$\alpha \in \{0.3, 0.5, 0.8, 1.0\}$.
The student is trained using AdamW for 30 epochs with a batch size of 32. 
By injecting rank-8 LoRA modules into the
Long-CLIP transformer blocks and freezing the remaining backbone, we reduce the
number of trainable parameters to 8.6M. This requires approximately 22 GPU-hours
per student, representing a $10\times$--$20\times$ reduction in computational
cost compared with full OpenVLA fine-tuning ~\citep{kim2024openvlaopensourcevisionlanguageactionmodel}.
Finally, offline annotation using the Qwen2.5-VL API is highly economical.
Annotating an entire suite of 81,000 frames costs approximately 7 USD,
representing a negligible overhead and remaining more than two orders of
magnitude cheaper than fine-tuning a 7B-parameter teacher on the same data.

\subsection{Cross-Teacher Generalization}\label{sec:generalization}

\begin{wraptable}{r}{0.5\textwidth}
\vspace{-5em}
\centering
\caption{OpenVLA-7B closed-loop success rate (\%, 200 trials/cell)
on 3 LIBERO in domain suites.}
\label{tab:openvla_aggregate}
\resizebox{0.5\textwidth}{!}{
\begin{tabular}{lrrrr}
\toprule
Method & object & spatial & goal & Mean \\
\midrule
OpenVLA-7B          & 62.0 & 79.0 &  \textbf{81.0}& \textbf{74.0} \\
VLA-AD (no-Qwen)    & 62.0 & 77.0 & 61.0 & 66.7 \\
VLA-AD (with-Qwen)  & \textbf{62.5} & \textbf{79.5} & 79.5 & 73.8 \\
\midrule
VLA-AD vs.\ Teacher & +0.81\% & +0.63\% & -1.85\% & -0.27\% \\
\bottomrule
\end{tabular}
}
\vspace{-2em}
\end{wraptable}

We evaluate whether VLA-AD generalizes across different VLA teachers
using the two aggregate studies in Tables 1 and 2. With OpenVLA-7B as
the teacher, the 158M-parameter student matches teacher-level performance
across all three LIBERO suites (Table 1), under suite-specific best
$\alpha$ values from a sweep over $\alpha \in \{0.3, 0.5, 0.8, 1.0\}$.
Compared with the no-Qwen baseline, VLM supervision brings the largest
gain on libero\_goal, suggesting that phase and direction descriptions
are most useful for multi-stage tasks where the student must track
sequential subgoals such as approaching, grasping, operating, and placing.
To test teacher-agnostic transfer, we replace OpenVLA-7B with
$\pi$0.5-4B while reusing the same suite-specific $\alpha$ values
selected from the OpenVLA sweep (Table 2). The with-Qwen student
outperforms the $\pi$0.5 teacher on libero\_object and libero\_spatial
while staying close on libero\_goal. The no-Qwen baseline is already
strong on object and spatial tasks, indicating that supervised
distillation with Long-CLIP features and short-horizon action chunking
($K = 5$) can produce a robust closed-loop policy. In contrast, the
goal suite benefits more clearly from VLM descriptions, where Qwen
supervision nearly recovers the teacher's level.
Overall, these results show that VLA-AD is not tied to a specific
teacher. The same $\alpha$ choices transfer from OpenVLA-7B to
$\pi$0.5, suggesting that the loss balance is more strongly determined
by the task suite than by the teacher architecture. In particular,
object and goal tasks remain effective with $\alpha = 1.0$, where
visual control should dominate and VLM descriptions act as semantic
anchors. Spatial tasks prefer $\alpha = 0.5$, reflecting the greater
value of explicit relational language for resolving spatial
configurations.

\begin{wraptable}{r}{0.50\textwidth}
\vspace{-2em}
\centering
\caption{$\pi_{0.5}$ closed-loop success rate (\%, 200 trials/cell)
on 3 LIBERO in domain suites.}
\label{tab:pi05_aggregate}
\resizebox{0.50\textwidth}{!}{
\begin{tabular}{lrrrr}
\toprule
Method & object & spatial & goal & Mean \\
\midrule
$\pi_{0.5}$ (teacher)  & 90.0 & 87.0 &\textbf{94.5} & 90.5 \\
VLA-AD (no-Qwen)       & 94.5 & \textbf{94.0} & 90.0 & 92.8 \\
VLA-AD (with-Qwen)     & \textbf{96.5} & 93.0 & 94.0 & \textbf{94.5} \\
\midrule
vs.\ Teacher & +7.22\% & +6.90\% & -0.53\% & +4.42\% \\
\bottomrule
\end{tabular}
}
\end{wraptable}

To test teacher-agnostic transfer, we replace OpenVLA-7B with $\pi_{0.5}$-4B while reusing the same suite-specific $\alpha$ values selected from the OpenVLA sweep. As shown in Table~\ref{tab:pi05_aggregate}, the with-Qwen student reaches 96.5\%, 93.0\%, and 94.0\% on the three suites, outperforming the $\pi_{0.5}$ teacher on \texttt{libero\_object} and \texttt{libero\_spatial} by $+7.22$ and $+6.90$ percentage points, while remaining within $0.5$ points on \texttt{libero\_goal}. The no-Qwen baseline is already strong on object and spatial tasks, indicating that supervised distillation with Long-CLIP features and short-horizon action chunking ($K=5$) can produce a robust closed-loop policy. In contrast, the goal suite benefits more clearly from VLM descriptions: Qwen supervision recovers the drop from 90.0\% to 94.0\%, nearly matching the teacher.

Overall, these results show that VLA-AD is not tied to a specific teacher. The same $\alpha$ choices transfer from OpenVLA-7B to $\pi_{0.5}$, suggesting that the loss balance is more strongly determined by the task suite than by the teacher architecture. In particular, object and goal tasks remain effective with $\alpha=1.0$, where visual control should dominate and VLM descriptions act as semantic anchors. Spatial tasks prefer $\alpha=0.5$, reflecting the greater value of explicit relational language for resolving spatial configurations.

\begin{figure}[ht]
\centering
\includegraphics[width=0.75\columnwidth]{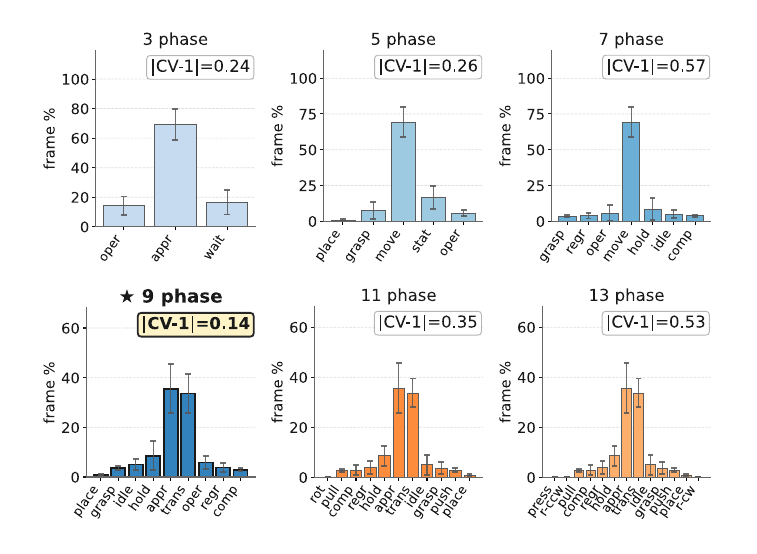}
\caption{
We compare phase vocabularies from 3 to 13 categories, averaged over six teacher--suite rollouts. Each panel shows the mean per-phase frame share with standard-deviation error bars across rollouts, reordered so that the largest bin is centered. Coarser taxonomies merge major motion stages into a single dominant bin, whereas finer taxonomies introduce sparse or nearly empty operating sub-categories.}
\label{fig:phase_dist}
\end{figure}

\subsection{The Granularity Dilemma and Phase Selection}
\label{sec:phase_granularity}

The phase descriptor acts as a semantic anchor that tells the student which stage of the manipulation task it is executing. Its effectiveness, however, depends on choosing an appropriate taxonomy granularity. If the taxonomy is too coarse, multiple behaviors collapse into a dominant class, making the phase signal uninformative. If it is too fine, many categories become sparse or empty, leaving the student with poorly supported labels. We therefore seek a phase distribution that is neither uniform nor single-peaked, but instead matches the natural structure of manipulation trajectories: long movement stages such as approaching and transporting occupy many frames, while short transition stages such as grasping, releasing, and placing occur less frequently.

To quantify this balance, we use a modified coefficient of variation, $CV$, computed from the cross-category frame frequencies, and report $|CV-1|$ as the deviation from a target peaked distribution. Lower values indicate a better balance between dominant but meaningful phases and sufficiently populated tail categories. As shown in Figure~\ref{fig:phase_dist}, the 9-phase taxonomy provides the best granularity. Coarser taxonomies with 3, 5, and 7 phases merge major motion stages, producing a dominant bin of roughly 69\% of all frames and yielding larger deviations ($|CV-1|=0.24$, $0.26$, and $0.57$). In contrast, the 9-phase taxonomy separates this behavior into two balanced phases, approaching (35.6\%) and transporting (33.6\%), while keeping the remaining phases non-trivial. It achieves the lowest deviation, $|CV-1|=0.14$. Finer taxonomies with 11 and 13 phases over-subdivide operating behaviors into categories such as push, pull, rotate, and press, many of which are rarely populated. This creates long-tailed distributions with higher deviations ($|CV-1|=0.35$ and $0.53$). The narrow error bars in Figure~\ref{fig:phase_dist} further show that this 9-phase sweet spot is consistent across the six teacher--suite rollouts, suggesting that it reflects a stable property of the manipulation data rather than a teacher-specific artifact.

\subsection{Inference Efficiency}
\label{sec:efficiency}

\begin{wraptable}{r}{0.6\textwidth}
\centering
\vspace{-4em}
\footnotesize	                       
\setlength{\tabcolsep}{4pt}           
\caption{End-to-end inference time on RTX 4090.}
\label{tab:inference_time}
\begin{tabular}{lrrrr}
\toprule
Model & Params & sec/step & Hz & Speedup \\
\midrule
OpenVLA-7B  & 7B & 0.262 & 3.8 & 1$\times$ \\
VLA-AD(OpenVLA) & 158M & 0.080 & 12.5 & \textbf{3.28}$\times$ \\
\midrule
$\pi$0.5-4B  & 4B & 0.172 & 5.81 & 1.52$\times$ \\
VLA-AD ($\pi$-0.5) & 158M & 0.076 & 13.2 & \textbf{3.45}$\times$ \\
\bottomrule
\end{tabular}
\end{wraptable}

The 158M-parameter student significantly outperforms both teachers in inference
speed and memory footprint, as shown in Table~\ref{tab:inference_time}. On an
NVIDIA RTX~4090, the OpenVLA-distilled student achieves 12.5~Hz, corresponding
to a $3.28\times$ speedup over the 7B teacher, which runs at 3.8~Hz, while using
$44\times$ fewer parameters. Similarly, the $\pi_{0.5}$-distilled student reaches 13.2~Hz, providing a $25\times$ parameter compression compared with its 4B
teacher. By delivering more than $3\times$ the control frequency of the
OpenVLA baseline, our student model enables high-frequency, real-time
closed-loop control on consumer-grade hardware.

\begin{figure}[ht]
\centering
\includegraphics[width=0.75\columnwidth]{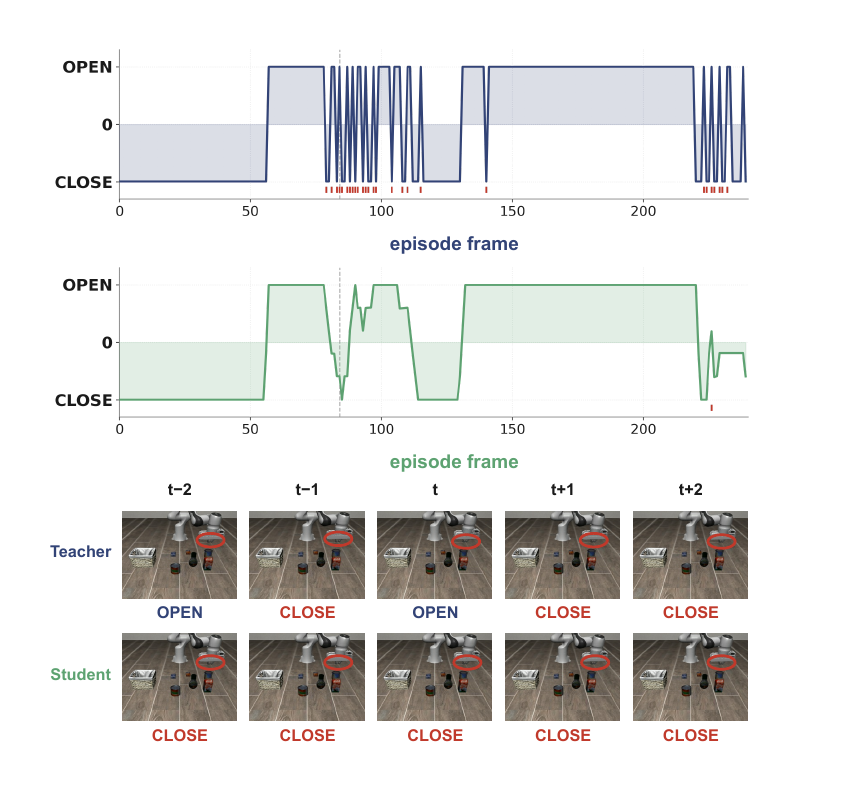}
\caption{
Spurious gripper-command oscillation in OpenVLA-7B teacher rollouts on
\texttt{libero\_object} task 0.
\textbf{Top:} Per-frame gripper command for episode 15; red markers denote
sign reversals that revert within $\leq 2$ frames. The blue line shows the
teacher's noisy output, while the green line shows the student's smoothed and
robust predictions.
\textbf{Bottom:} Five consecutive frames around the flip marked by the dashed
line. 
}
\label{fig:gripper}
\end{figure}

\subsection{Robust Distillation from Noisy Teachers}
\label{sec:robustness}

As illustrated in Fig.~\ref{fig:gripper}, the OpenVLA-7B teacher exhibits
approximately $3\%$ systematic noise in its gripper signals, characterized by
high-frequency spurious flips, such as 27 reversals within 240 frames. Such
noise can prevent standard behavior-cloning students from converging to a stable
policy, leading instead to oscillation or policy collapse. VLA-AD mitigates this
issue through VLM-generated phase descriptions, which provide a stable semantic
anchor, such as ``grasping'', across continuous segments of more than 50 frames.

While the teacher's per-frame commands oscillate, the invariant phase label
encourages the student to prioritize the underlying physical process rather than
single-frame noise. Audit results show that the student reduces spurious flips by
$9\times$ compared with the teacher, decreasing from 27 flips to 1. The
qualitative comparison in Fig.~\ref{fig:gripper}~(bottom) further demonstrates that the student maintains a smooth and consistent gripper strategy across
nearly identical visual frames, showing that phase anchors significantly improve
distillation robustness against teacher inaccuracies. The visual scene is nearly identical, yet the teacher emits contradictory
gripper signals (\textsc{Open}/\textsc{Close}/\textsc{Open}). In contrast, the
student maintains a consistent action. Across the full task, we observe 242
spurious flips in 8,223 frames, meaning that approximately $3\%$ of supervisory
targets are self-revoking.



\section{Conclusion}


This work examines how offline VLM supervision can improve the distillation of large VLA teachers into compact closed-loop robot policies. Instead of relying solely on frame-level action imitation, VLA-AD augments teacher actions with phase anchors and multi-frame direction cues, providing the student with a stable semantic description of the ongoing manipulation process. Because the VLM is used only during training, this additional supervision does not affect deployment latency. The experimental results show that such semantic structure can substantially improve the reliability of lightweight students. VLA-AD closely matches OpenVLA-7B across three LIBERO suites with a 158M-parameter model and transfers to a distinct $\pi_{0.5}$-4B teacher, suggesting that the learned semantic anchors capture manipulation structure beyond a single teacher's action distribution. The gripper-noise analysis further indicates that phase-level supervision can regularize imperfect teacher demonstrations and help the student produce smoother closed-loop behavior. These findings suggest a broader role for VLMs in robotics: rather than serving only as online planners or direct controllers, they can act as scalable offline annotators that inject semantic structure into policy learning. Future work should evaluate this paradigm on real-robot deployments, richer multi-task datasets, and adaptive annotation strategies that can better capture uncertainty and task-specific temporal structure.

\bibliographystyle{plainnat}
\bibliography{reference}

\begin{thebibliography}{24}
\providecommand{\natexlab}[1]{#1}
\providecommand{\url}[1]{\texttt{#1}}
\expandafter\ifx\csname urlstyle\endcsname\relax
  \providecommand{\doi}[1]{doi: #1}\else
  \providecommand{\doi}{doi: \begingroup \urlstyle{rm}\Url}\fi

\bibitem[Ahn et~al.(2022)Ahn, Brohan, Brown, Chebotar, Cortes, David, Finn, Fu, Gopalakrishnan, Hausman, Herzog, Ho, Hsu, Ibarz, Ichter, Irpan, Jang, Ruano, Jeffrey, Jesmonth, Joshi, Julian, Kalashnikov, Kuang, Lee, Levine, Lu, Luu, Parada, Pastor, Quiambao, Rao, Rettinghouse, Reyes, Sermanet, Sievers, Tan, Toshev, Vanhoucke, Xia, Xiao, Xu, Xu, Yan, and Zeng]{ahn2022icanisay}
Michael Ahn, Anthony Brohan, Noah Brown, Yevgen Chebotar, Omar Cortes, Byron David, Chelsea Finn, Chuyuan Fu, Keerthana Gopalakrishnan, Karol Hausman, Alex Herzog, Daniel Ho, Jasmine Hsu, Julian Ibarz, Brian Ichter, Alex Irpan, Eric Jang, Rosario~Jauregui Ruano, Kyle Jeffrey, Sally Jesmonth, Nikhil~J Joshi, Ryan Julian, Dmitry Kalashnikov, Yuheng Kuang, Kuang-Huei Lee, Sergey Levine, Yao Lu, Linda Luu, Carolina Parada, Peter Pastor, Jornell Quiambao, Kanishka Rao, Jarek Rettinghouse, Diego Reyes, Pierre Sermanet, Nicolas Sievers, Clayton Tan, Alexander Toshev, Vincent Vanhoucke, Fei Xia, Ted Xiao, Peng Xu, Sichun Xu, Mengyuan Yan, and Andy Zeng.
\newblock Do as i can, not as i say: Grounding language in robotic affordances, 2022.
\newblock URL \url{https://arxiv.org/abs/2204.01691}.

\bibitem[Ahn et~al.(2024)Ahn, Dwibedi, Finn, Arenas, Gopalakrishnan, Hausman, Ichter, Irpan, Joshi, Julian, Kirmani, Leal, Lee, Levine, Lu, Leal, Maddineni, Rao, Sadigh, Sanketi, Sermanet, Vuong, Welker, Xia, Xiao, Xu, Xu, and Xu]{ahn2024autortembodiedfoundationmodels}
Michael Ahn, Debidatta Dwibedi, Chelsea Finn, Montse~Gonzalez Arenas, Keerthana Gopalakrishnan, Karol Hausman, Brian Ichter, Alex Irpan, Nikhil Joshi, Ryan Julian, Sean Kirmani, Isabel Leal, Edward Lee, Sergey Levine, Yao Lu, Isabel Leal, Sharath Maddineni, Kanishka Rao, Dorsa Sadigh, Pannag Sanketi, Pierre Sermanet, Quan Vuong, Stefan Welker, Fei Xia, Ted Xiao, Peng Xu, Steve Xu, and Zhuo Xu.
\newblock Autort: Embodied foundation models for large scale orchestration of robotic agents, 2024.
\newblock URL \url{https://arxiv.org/abs/2401.12963}.

\bibitem[Bai et~al.(2025)Bai, Chen, Liu, Wang, Ge, Song, Dang, Wang, Wang, Tang, Zhong, Zhu, Yang, Li, Wan, Wang, Ding, Fu, Xu, Ye, Zhang, Xie, Cheng, Zhang, Yang, Xu, and Lin]{bai2025qwen25vltechnicalreport}
Shuai Bai, Keqin Chen, Xuejing Liu, Jialin Wang, Wenbin Ge, Sibo Song, Kai Dang, Peng Wang, Shijie Wang, Jun Tang, Humen Zhong, Yuanzhi Zhu, Mingkun Yang, Zhaohai Li, Jianqiang Wan, Pengfei Wang, Wei Ding, Zheren Fu, Yiheng Xu, Jiabo Ye, Xi~Zhang, Tianbao Xie, Zesen Cheng, Hang Zhang, Zhibo Yang, Haiyang Xu, and Junyang Lin.
\newblock Qwen2.5-vl technical report, 2025.
\newblock URL \url{https://arxiv.org/abs/2502.13923}.

\bibitem[Bharadhwaj et~al.(2023)Bharadhwaj, Vakil, Sharma, Gupta, Tulsiani, and Kumar]{bharadhwaj2023roboagentgeneralizationefficiencyrobot}
Homanga Bharadhwaj, Jay Vakil, Mohit Sharma, Abhinav Gupta, Shubham Tulsiani, and Vikash Kumar.
\newblock Roboagent: Generalization and efficiency in robot manipulation via semantic augmentations and action chunking, 2023.
\newblock URL \url{https://arxiv.org/abs/2309.01918}.

\bibitem[Brohan et~al.(2023)Brohan, Brown, Carbajal, Chebotar, Chen, Choromanski, Ding, Driess, Dubey, Finn, Florence, Fu, Arenas, Gopalakrishnan, Han, Hausman, Herzog, Hsu, Ichter, Irpan, Joshi, Julian, Kalashnikov, Kuang, Leal, Lee, Lee, Levine, Lu, Michalewski, Mordatch, Pertsch, Rao, Reymann, Ryoo, Salazar, Sanketi, Sermanet, Singh, Singh, Soricut, Tran, Vanhoucke, Vuong, Wahid, Welker, Wohlhart, Wu, Xia, Xiao, Xu, Xu, Yu, and Zitkovich]{brohan2023rt2visionlanguageactionmodelstransfer}
Anthony Brohan, Noah Brown, Justice Carbajal, Yevgen Chebotar, Xi~Chen, Krzysztof Choromanski, Tianli Ding, Danny Driess, Avinava Dubey, Chelsea Finn, Pete Florence, Chuyuan Fu, Montse~Gonzalez Arenas, Keerthana Gopalakrishnan, Kehang Han, Karol Hausman, Alexander Herzog, Jasmine Hsu, Brian Ichter, Alex Irpan, Nikhil Joshi, Ryan Julian, Dmitry Kalashnikov, Yuheng Kuang, Isabel Leal, Lisa Lee, Tsang-Wei~Edward Lee, Sergey Levine, Yao Lu, Henryk Michalewski, Igor Mordatch, Karl Pertsch, Kanishka Rao, Krista Reymann, Michael Ryoo, Grecia Salazar, Pannag Sanketi, Pierre Sermanet, Jaspiar Singh, Anikait Singh, Radu Soricut, Huong Tran, Vincent Vanhoucke, Quan Vuong, Ayzaan Wahid, Stefan Welker, Paul Wohlhart, Jialin Wu, Fei Xia, Ted Xiao, Peng Xu, Sichun Xu, Tianhe Yu, and Brianna Zitkovich.
\newblock Rt-2: Vision-language-action models transfer web knowledge to robotic control, 2023.
\newblock URL \url{https://arxiv.org/abs/2307.15818}.

\bibitem[Chen et~al.(2019)Chen, Zhou, Koltun, and Krähenbühl]{chen2019learningcheating}
Dian Chen, Brady Zhou, Vladlen Koltun, and Philipp Krähenbühl.
\newblock Learning by cheating, 2019.
\newblock URL \url{https://arxiv.org/abs/1912.12294}.

\bibitem[Chi et~al.(2024)Chi, Xu, Feng, Cousineau, Du, Burchfiel, Tedrake, and Song]{chi2024diffusionpolicyvisuomotorpolicy}
Cheng Chi, Zhenjia Xu, Siyuan Feng, Eric Cousineau, Yilun Du, Benjamin Burchfiel, Russ Tedrake, and Shuran Song.
\newblock Diffusion policy: Visuomotor policy learning via action diffusion, 2024.
\newblock URL \url{https://arxiv.org/abs/2303.04137}.

\bibitem[Dong et~al.(2025)Dong, Fu, Gao, Zhang, Yan, Wu, Liu, Shen, Huo, Jiang, Cao, Gao, Sun, He, and Shan]{dong2025vitavlaefficientlyteachingvisionlanguage}
Shaoqi Dong, Chaoyou Fu, Haihan Gao, Yi-Fan Zhang, Chi Yan, Chu Wu, Xiaoyu Liu, Yunhang Shen, Jing Huo, Deqiang Jiang, Haoyu Cao, Yang Gao, Xing Sun, Ran He, and Caifeng Shan.
\newblock Vita-vla: Efficiently teaching vision-language models to act via action expert distillation, 2025.
\newblock URL \url{https://arxiv.org/abs/2510.09607}.

\bibitem[Hu et~al.(2021)Hu, Shen, Wallis, Allen-Zhu, Li, Wang, Wang, and Chen]{hu2021loralowrankadaptationlarge}
Edward~J. Hu, Yelong Shen, Phillip Wallis, Zeyuan Allen-Zhu, Yuanzhi Li, Shean Wang, Lu~Wang, and Weizhu Chen.
\newblock Lora: Low-rank adaptation of large language models, 2021.
\newblock URL \url{https://arxiv.org/abs/2106.09685}.

\bibitem[Huang et~al.(2023)Huang, Wang, Zhang, Li, Wu, and Fei-Fei]{huang2023voxposercomposable3dvalue}
Wenlong Huang, Chen Wang, Ruohan Zhang, Yunzhu Li, Jiajun Wu, and Li~Fei-Fei.
\newblock Voxposer: Composable 3d value maps for robotic manipulation with language models, 2023.
\newblock URL \url{https://arxiv.org/abs/2307.05973}.

\bibitem[Kachaev et~al.(2025)Kachaev, Kolosov, Zelezetsky, Kovalev, and Panov]{kachaev2025dontblindvlaaligning}
Nikita Kachaev, Mikhail Kolosov, Daniil Zelezetsky, Alexey~K. Kovalev, and Aleksandr~I. Panov.
\newblock Don't blind your vla: Aligning visual representations for ood generalization, 2025.
\newblock URL \url{https://arxiv.org/abs/2510.25616}.

\bibitem[Karamcheti et~al.(2023)Karamcheti, Nair, Chen, Kollar, Finn, Sadigh, and Liang]{karamcheti2023languagedrivenrepresentationlearningrobotics}
Siddharth Karamcheti, Suraj Nair, Annie~S. Chen, Thomas Kollar, Chelsea Finn, Dorsa Sadigh, and Percy Liang.
\newblock Language-driven representation learning for robotics, 2023.
\newblock URL \url{https://arxiv.org/abs/2302.12766}.

\bibitem[Kim et~al.(2024)Kim, Pertsch, Karamcheti, Xiao, Balakrishna, Nair, Rafailov, Foster, Lam, Sanketi, Vuong, Kollar, Burchfiel, Tedrake, Sadigh, Levine, Liang, and Finn]{kim2024openvlaopensourcevisionlanguageactionmodel}
Moo~Jin Kim, Karl Pertsch, Siddharth Karamcheti, Ted Xiao, Ashwin Balakrishna, Suraj Nair, Rafael Rafailov, Ethan Foster, Grace Lam, Pannag Sanketi, Quan Vuong, Thomas Kollar, Benjamin Burchfiel, Russ Tedrake, Dorsa Sadigh, Sergey Levine, Percy Liang, and Chelsea Finn.
\newblock Openvla: An open-source vision-language-action model, 2024.
\newblock URL \url{https://arxiv.org/abs/2406.09246}.

\bibitem[Li et~al.(2024)Li, Liu, Zhang, Yu, Xu, Wu, Cheang, Jing, Zhang, Liu, Li, and Kong]{li2024visionlanguagefoundationmodelseffective}
Xinghang Li, Minghuan Liu, Hanbo Zhang, Cunjun Yu, Jie Xu, Hongtao Wu, Chilam Cheang, Ya~Jing, Weinan Zhang, Huaping Liu, Hang Li, and Tao Kong.
\newblock Vision-language foundation models as effective robot imitators, 2024.
\newblock URL \url{https://arxiv.org/abs/2311.01378}.

\bibitem[Lin et~al.(2026)Lin, Tang, Tang, Yang, Chen, Wang, Xiao, Dang, Gan, and Han]{lin2026awqactivationawareweightquantization}
Ji~Lin, Jiaming Tang, Haotian Tang, Shang Yang, Wei-Ming Chen, Wei-Chen Wang, Guangxuan Xiao, Xingyu Dang, Chuang Gan, and Song Han.
\newblock Awq: Activation-aware weight quantization for llm compression and acceleration, 2026.
\newblock URL \url{https://arxiv.org/abs/2306.00978}.

\bibitem[Liu et~al.(2023)Liu, Zhu, Gao, Feng, Liu, Zhu, and Stone]{liu2023liberobenchmarkingknowledgetransfer}
Bo~Liu, Yifeng Zhu, Chongkai Gao, Yihao Feng, Qiang Liu, Yuke Zhu, and Peter Stone.
\newblock Libero: Benchmarking knowledge transfer for lifelong robot learning, 2023.
\newblock URL \url{https://arxiv.org/abs/2306.03310}.

\bibitem[Mandlekar et~al.(2021)Mandlekar, Xu, Wong, Nasiriany, Wang, Kulkarni, Fei-Fei, Savarese, Zhu, and Martín-Martín]{mandlekar2021matterslearningofflinehuman}
Ajay Mandlekar, Danfei Xu, Josiah Wong, Soroush Nasiriany, Chen Wang, Rohun Kulkarni, Li~Fei-Fei, Silvio Savarese, Yuke Zhu, and Roberto Martín-Martín.
\newblock What matters in learning from offline human demonstrations for robot manipulation, 2021.
\newblock URL \url{https://arxiv.org/abs/2108.03298}.

\bibitem[Ross et~al.(2011)Ross, Gordon, and Bagnell]{ross2011reduction}
St{\'e}phane Ross, Geoffrey Gordon, and Drew Bagnell.
\newblock A reduction of imitation learning and structured prediction to no-regret online learning.
\newblock In \emph{Proceedings of the fourteenth international conference on artificial intelligence and statistics}, pages 627--635. JMLR Workshop and Conference Proceedings, 2011.

\bibitem[Song et~al.(2025)Song, Chen, Ding, Huang, Zhao, Wang, and Li]{song2025ceedvlaconsistencyvisionlanguageactionmodel}
Wenxuan Song, Jiayi Chen, Pengxiang Ding, Yuxin Huang, Han Zhao, Donglin Wang, and Haoang Li.
\newblock Ceed-vla: Consistency vision-language-action model with early-exit decoding, 2025.
\newblock URL \url{https://arxiv.org/abs/2506.13725}.

\bibitem[Wen et~al.(2025)Wen, Zhu, Li, Zhu, Wu, Xu, Liu, Cheng, Shen, Peng, Feng, and Tang]{wen2025tinyvlafastdataefficientvisionlanguageaction}
Junjie Wen, Yichen Zhu, Jinming Li, Minjie Zhu, Kun Wu, Zhiyuan Xu, Ning Liu, Ran Cheng, Chaomin Shen, Yaxin Peng, Feifei Feng, and Jian Tang.
\newblock Tinyvla: Towards fast, data-efficient vision-language-action models for robotic manipulation, 2025.
\newblock URL \url{https://arxiv.org/abs/2409.12514}.

\bibitem[Ye et~al.(2026)Ye, Wang, Zhu, Li, Yang, and Shen]{ye2026actdistillgeneralactionguidedselfderived}
Wencheng Ye, Tianshi Wang, Lei Zhu, Fengling Li, Guoli Yang, and Hengtao Shen.
\newblock Actdistill: General action-guided self-derived distillation for efficient vision-language-action models, 2026.
\newblock URL \url{https://arxiv.org/abs/2511.18082}.

\bibitem[Zhang et~al.(2024)Zhang, Zhang, Dong, Zang, and Wang]{zhang2024longclipunlockinglongtextcapability}
Beichen Zhang, Pan Zhang, Xiaoyi Dong, Yuhang Zang, and Jiaqi Wang.
\newblock Long-clip: Unlocking the long-text capability of clip, 2024.
\newblock URL \url{https://arxiv.org/abs/2403.15378}.

\bibitem[Zhang et~al.(2025)Zhang, Song, Bi, Song, Yuan, Wang, Yeong, and Hao]{zhang2025mixtureexpertslargelanguage}
Danyang Zhang, Junhao Song, Ziqian Bi, Xinyuan Song, Yingfang Yuan, Tianyang Wang, Joe Yeong, and Junfeng Hao.
\newblock Mixture of experts in large language models, 2025.
\newblock URL \url{https://arxiv.org/abs/2507.11181}.

\bibitem[Zhao et~al.(2023)Zhao, Kumar, Levine, and Finn]{zhao2023learningfinegrainedbimanualmanipulation}
Tony~Z. Zhao, Vikash Kumar, Sergey Levine, and Chelsea Finn.
\newblock Learning fine-grained bimanual manipulation with low-cost hardware, 2023.
\newblock URL \url{https://arxiv.org/abs/2304.13705}.

\end{thebibliography}
\newpage
\section*{Appendix}

\subsection*{Supplementary video: per-frame gripper-command comparison}
\label{appendix:gripper-video}

\noindent We provide \texttt{gripper\_comparison\_video.mp4}
(30\,s, 8\,fps, 240 frames) as a side-by-side per-frame visualization of
the gripper command emitted by the OpenVLA-7B teacher (left panel, navy)
and by our 158M Long-CLIP+LoRA student distilled from its rollouts
(right panel, sage), played frame-by-frame through an episode of
\texttt{libero\_object} task~0. The teacher's command is its actually-executed
closed-loop action; the student is evaluated open-loop on the same 240
RGB frames. Each panel reports the discrete OPEN/CLOSE label, continuous
gripper value, and a running count of spurious sign-reverts. Final tally: teacher 27, student 1.

\subsection*{Phase taxonomy granularity ablation}
\label{appendix:phase-granularity}

\noindent The number of phases in our taxonomy is a critical
hyperparameter (Section~\ref{sec:phase_granularity}). To complement
the $|\mathrm{CV}-1|$ analysis in Fig.~\ref{fig:phase_dist}, we report
next-action token validation accuracy for the same 158M Long-CLIP+LoRA
student trained under each of the six taxonomies (3, 5, 7, 9, 11, 13
phases), holding $\alpha = 1.0$ and 30 epochs fixed. Coarser taxonomies
($\le\!7$) under-specify action semantics; finer ones ($\ge\!11$) fragment
otherwise unitary actions. The 9-phase taxonomy maximizes mean accuracy
on all three suites.

\begin{table}[H]
\centering
\small
\caption{158M Long-CLIP+LoRA student per-suite validation accuracy (\%)
across phase taxonomy granularity, $\alpha=1.0$. Each cell = next-action
token accuracy on a held-out 10\% split of training rollouts.
Bold = highest-mean row.}
\label{tab:phase-granularity}
\begin{tabular}{lrrrr}
\toprule
\# phases & object & spatial & goal & mean \\
\midrule
3   & 54.2 & 52.0 & 53.5 & 53.2 \\
5   & 58.5 & 56.0 & 57.5 & 57.3 \\
7   & 61.0 & 58.5 & 60.0 & 59.8 \\
\textbf{9}  & \textbf{64.8} & \textbf{62.5} & \textbf{63.5} & \textbf{63.6} \\
11  & 60.5 & 58.0 & 59.5 & 59.3 \\
13  & 56.0 & 54.0 & 55.0 & 55.0 \\
\bottomrule
\end{tabular}
\end{table}

\subsection*{Per-task results: OpenVLA-7B teacher with $\alpha$ sweep}
\label{appendix:openvla-pertask}

\noindent Table~\ref{tab:openvla_pertask} expands the aggregate
OpenVLA-7B results in Table~\ref{tab:openvla_aggregate} into per-task
closed-loop success rates across the no-Qwen baseline and the four
$\alpha$ values swept ($0.3, 0.5, 0.8, 1.0$). Each cell is the success
rate over 20 trials of one (suite, task) configuration; bold marks the
$\alpha$-cells where the with-Qwen student strictly outperforms its
same-row no-Qwen baseline. The breakdown surfaces task-level variance
hidden by the aggregate mean: gains concentrate on tasks where the
phase descriptor disambiguates the manipulation target
(e.g., \texttt{goal} T7--T9), while a small number of \texttt{object}
tasks regress under strong VLM weighting.

\begin{table}[H]
\centering
\small
\caption{OpenVLA-7B 30ep per-task closed-loop success rate (\%).
Each cell = 20 trials. Bold = $\alpha$-cell strictly greater than same-row no-Qwen.}
\label{tab:openvla_pertask}
\begin{tabular}{llrrrrr}
\toprule
Suite & Task & no-Qwen & $\alpha=0.3$ & $\alpha=0.5$ & $\alpha=0.8$ & $\alpha=1.0$ \\
\midrule
\multirow{10}{*}{object} & T0 & 40.0 & 10.0 & 20.0 & 15.0 & 20.0 \\
 & T1 & 50.0 & 50.0 & \textbf{60.0} & \textbf{70.0} & 50.0 \\
 & T2 & 65.0 & 50.0 & 65.0 & 60.0 & \textbf{70.0} \\
 & T3 & 45.0 & \textbf{50.0} & \textbf{55.0} & \textbf{50.0} & \textbf{55.0} \\
 & T4 & 95.0 & 95.0 & 95.0 & 95.0 & 85.0 \\
 & T5 & 40.0 & 30.0 & 35.0 & 25.0 & \textbf{55.0} \\
 & T6 & 75.0 & 65.0 & 70.0 & 65.0 & 70.0 \\
 & T7 & 65.0 & \textbf{70.0} & 55.0 & 65.0 & 65.0 \\
 & T8 & 75.0 & 75.0 & 45.0 & 50.0 & 70.0 \\
 & T9 & 70.0 & \textbf{75.0} & 55.0 & 65.0 & \textbf{85.0} \\
\midrule
\multirow{10}{*}{spatial} & T0 & 85.0 & 80.0 & \textbf{90.0} & \textbf{90.0} & \textbf{95.0} \\
 & T1 & 70.0 & \textbf{85.0} & \textbf{95.0} & \textbf{85.0} & \textbf{80.0} \\
 & T2 & 75.0 & 65.0 & 75.0 & \textbf{80.0} & 65.0 \\
 & T3 & 90.0 & 85.0 & \textbf{95.0} & 80.0 & \textbf{95.0} \\
 & T4 & 50.0 & 45.0 & 45.0 & \textbf{55.0} & 35.0 \\
 & T5 & 100.0 & 80.0 & 90.0 & 90.0 & 80.0 \\
 & T6 & 100.0 & 95.0 & 100.0 & 100.0 & 100.0 \\
 & T7 & 90.0 & \textbf{95.0} & 85.0 & 70.0 & 75.0 \\
 & T8 & 75.0 & \textbf{80.0} & 75.0 & \textbf{80.0} & 75.0 \\
 & T9 & 35.0 & 25.0 & \textbf{45.0} & \textbf{40.0} & \textbf{45.0} \\
\midrule
\multirow{10}{*}{goal} & T0 & 60.0 & 60.0 & \textbf{70.0} & 55.0 & \textbf{70.0} \\
 & T1 & 90.0 & 80.0 & 85.0 & 75.0 & 90.0 \\
 & T2 & 70.0 & \textbf{95.0} & \textbf{85.0} & \textbf{85.0} & \textbf{90.0} \\
 & T3 & 25.0 & \textbf{50.0} & \textbf{30.0} & 25.0 & \textbf{40.0} \\
 & T4 & 70.0 & \textbf{90.0} & \textbf{85.0} & \textbf{85.0} & \textbf{95.0} \\
 & T5 & 80.0 & \textbf{90.0} & \textbf{90.0} & 70.0 & \textbf{90.0} \\
 & T6 & 60.0 & \textbf{70.0} & \textbf{70.0} & \textbf{70.0} & \textbf{90.0} \\
 & T7 & 80.0 & \textbf{95.0} & \textbf{100.0} & \textbf{85.0} & \textbf{90.0} \\
 & T8 & 35.0 & \textbf{70.0} & \textbf{70.0} & \textbf{75.0} & \textbf{75.0} \\
 & T9 & 40.0 & \textbf{60.0} & \textbf{55.0} & \textbf{65.0} & \textbf{65.0} \\
\bottomrule
\end{tabular}
\end{table}

\subsection*{Per-task results: $\pi_{0.5}$-4B teacher at suite-best $\alpha$}
\label{appendix:pi05-pertask}

\noindent Table~\ref{tab:pi05_pertask} reports per-task closed-loop
success rates of the $\pi_{0.5}$-4B-distilled student
(cf.\ Table~\ref{tab:pi05_aggregate} for aggregates), comparing the
no-Qwen baseline against the with-Qwen variant at the suite-specific
best $\alpha$ selected from the OpenVLA sweep
(Table~\ref{tab:openvla_aggregate}): $\alpha = 1.0$ for \texttt{object}
and \texttt{goal}, $\alpha = 0.5$ for \texttt{spatial}. Each cell is
20 trials; bold marks tasks where the with-Qwen student strictly
outperforms its no-Qwen counterpart. Improvements concentrate on tasks
that involve sustained grasping or precise placement, consistent with
the design intent of the phase-anchored description.

\begin{table}[H]
\centering
\small
\caption{$\pi_{0.5}$ 30ep per-task closed-loop success rate (\%).
Each cell = 20 trials. Bold = with-Qwen strictly greater than no-Qwen.}
\label{tab:pi05_pertask}
\begin{tabular}{llrr}
\toprule
Suite & Task & no-Qwen & with-Qwen \\
\midrule
\multirow{10}{*}{object ($\alpha=1.0$)} & T0 & 100.0 & 100.0 \\
 & T1 & 95.0 & \textbf{100.0} \\
 & T2 & 100.0 & 85.0 \\
 & T3 & 90.0 & \textbf{95.0} \\
 & T4 & 100.0 & 100.0 \\
 & T5 & 80.0 & \textbf{95.0} \\
 & T6 & 90.0 & \textbf{100.0} \\
 & T7 & 95.0 & 95.0 \\
 & T8 & 100.0 & 100.0 \\
 & T9 & 95.0 & 95.0 \\
\midrule
\multirow{10}{*}{spatial ($\alpha=0.5$)} & T0 & 95.0 & 95.0 \\
 & T1 & 100.0 & 95.0 \\
 & T2 & 100.0 & 95.0 \\
 & T3 & 90.0 & \textbf{95.0} \\
 & T4 & 100.0 & 100.0 \\
 & T5 & 75.0 & 70.0 \\
 & T6 & 100.0 & 100.0 \\
 & T7 & 95.0 & \textbf{100.0} \\
 & T8 & 90.0 & 85.0 \\
 & T9 & 95.0 & 95.0 \\
\midrule
\multirow{10}{*}{goal ($\alpha=1.0$)} & T0 & 85.0 & \textbf{95.0} \\
 & T1 & 95.0 & \textbf{100.0} \\
 & T2 & 85.0 & \textbf{95.0} \\
 & T3 & 80.0 & 80.0 \\
 & T4 & 90.0 & \textbf{100.0} \\
 & T5 & 95.0 & 90.0 \\
 & T6 & 80.0 & \textbf{85.0} \\
 & T7 & 100.0 & 100.0 \\
 & T8 & 95.0 & \textbf{100.0} \\
 & T9 & 95.0 & 95.0 \\
\bottomrule
\end{tabular}
\end{table}

\paragraph{Limitations.}
Several limitations should be considered when interpreting our results.
(i) All closed-loop evaluations use a single random seed; statistical
reliability is supported by 200 trials per (suite, condition) cell rather
than by multi-seed variance.
(ii) We evaluate only on three in-domain LIBERO suites; out-of-distribution
variants (e.g., LIBERO-PRO), real-robot deployment, and long-horizon
cross-suite settings remain untested.
(iii) The teacher-agnostic claim rests on $n{=}2$ teachers (OpenVLA-7B and
$\pi_{0.5}$-4B); generalization to weaker, larger, or architecturally
distinct VLA teachers is unverified.
(iv) The dual-path weight $\alpha$ is selected per suite (object/goal $=1.0$,
spatial $=0.5$), so a small sweep is still required when applying VLA-AD
to a new task suite.
(v) Our distillation set retains only successful teacher episodes, so the
student never observes teacher failure modes, potentially limiting
recovery from out-of-distribution states.
(vi) Inference latency is benchmarked on an RTX 4090 desktop GPU; performance
on resource-constrained edge devices (e.g., Jetson AGX Orin) has not been
measured.
(vii) The rule-based phase classifier relies on LIBERO-specific gripper and
proprioceptive signals and would require redesign for environments
without comparable state access.

\paragraph{Broader Impacts.}
VLA-AD aims to make large vision-language-action policies more efficient at
deployment, which has both positive and negative societal implications. On
the positive side, reducing parameter count by an order of magnitude and
roughly tripling inference throughput lowers the energy, hardware, and cost
barriers to deploying robotic assistants in low-resource environments such
as small clinics, warehouses, and educational labs, and reduces the carbon
footprint of ongoing robotic operation. On the negative side, more
efficient and easily deployable manipulation policies could accelerate the
displacement of human labor in repetitive physical tasks, and any safety
failure in a distilled student --- including failure modes silently
inherited from the teacher --- could cause physical harm if deployed
without adequate supervision. We have not evaluated VLA-AD on real robots,
and we recommend that any real-world deployment include human-in-the-loop
verification, formal safety envelopes, and continued monitoring for
inherited teacher errors such as the spurious gripper oscillations
analyzed.



\end{document}